\pdfoutput=1
\documentclass[11pt]{article}

\usepackage[preprint]{acl}
\usepackage{times}
\usepackage{latexsym}
\usepackage[T1]{fontenc}
\usepackage[utf8]{inputenc}
\usepackage{microtype}
\usepackage{inconsolata}
\usepackage{graphicx}
\usepackage{amsmath}
\usepackage{multirow}
\usepackage{booktabs} 
\usepackage{cleveref}
\usepackage{colortbl}
\usepackage{listings}
\definecolor{light-light-gray}{gray}{0.92}

\title{Data or Language Supervision: What Makes CLIP Better than DINO?}

\author{Yiming Liu$^{{1,2},\star}$, Yuhui Zhang$^{{1},\star}$, Dhruba Ghosh$^{1}$, \\ \textbf{Ludwig Schmidt}$^{{1},\dagger}$\textbf{,} \textbf{Serena Yeung-Levy}$^{{1},\dagger}$ \\ $^{1}$Stanford University, $^{2}$Tsinghua University \\ $^\star$equal contribution, $^\dagger$equal advising \\ \texttt{\{ymingliu, yuhuiz, djghosh, ludwigsc, syyeung\}@stanford.edu} }

\begin{document}

\maketitle
\begin{abstract}

CLIP outperforms self-supervised models like DINO as vision encoders for vision-language models (VLMs), but it remains unclear whether this advantage stems from CLIP’s language supervision or its much larger training data. To disentangle these factors, we pre-train CLIP and DINO under controlled settings—using the same architecture, dataset, and training configuration—achieving similar ImageNet accuracy. Embedding analysis shows that CLIP captures high-level semantics (e.g., object categories, text), while DINO is more responsive to low-level features like colors and styles. When integrated into VLMs and evaluated on 20 VQA benchmarks, CLIP excels at text-intensive tasks, while DINO slightly outperforms on vision-centric ones. Variants of language supervision (e.g., sigmoid loss, pre-trained language encoders) yield limited gains. Our findings provide scientific insights into vision encoder design and its impact on VLM performance.\footnote{Code available at \url{https://github.com/leo1oel/Controlled-CLIP-DINO}.}

\end{abstract}

\section{Introduction}
\label{sec:intro}

Vision-language models, such as GPT-4o \cite{openaicom2025gpt04vision0} and Claude \cite{anthropiccom2025introducing}, have demonstrated transformative capabilities in interpreting and reasoning over visual inputs. These models typically consist of a vision encoder, a language model, and a connector module bridging the two~\cite{Liu2023VisualIT}. The vision encoder—serving as the “eyes” of the system—plays a critical role in transmitting visual information to the language model, which acts as the “brain” that interprets it.

\begin{figure}[!tb]
    \centering
    \includegraphics[width=\linewidth]{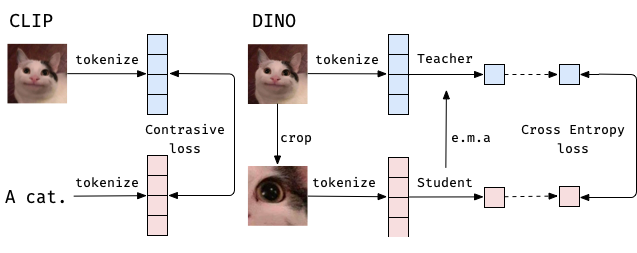}
    \small
    \vspace{1em}
    \rowcolors{2}{white}{light-light-gray}
\setlength\tabcolsep{5pt}
\renewcommand{\arraystretch}{1.1}
    \begin{tabular}{lcc}
    \toprule
    \textbf{Model} & \textbf{Data} $\times$ \textbf{Epochs} & \textbf{V100 GPU Hours} \\
    \midrule
    CLIP & 400M$\times$32 (12.8B) & 73K \\
    SigLIP & 40B & >100K \\
    DINO & 1.28M$\times$300 (384M) & 1K \\
    \bottomrule
    \end{tabular}
    \vspace{-1.3em}
    \caption{CLIP and DINO represent two predominant paradigms of vision encoders, differing in two key aspects: (1) CLIP is trained with language supervision, whereas DINO uses image-only self-supervision; (2) CLIP and its variant SigLIP are trained on datasets that are up to 100 times larger than those used for DINO. These differences make it difficult to disentangle whether CLIP’s superior performance in vision-language models stems from its training objective or the scale of its training data.}
    \vspace{-1em}
    \label{fig:clip}
\end{figure}

Recent studies have shown that CLIP \cite{DBLP:conf/icml/RadfordKHRGASAM21}, particularly its variant SigLIP \cite{zhai2023sigmoid}, has emerged as the most effective vision encoder for building VLMs, significantly outperforming DINO-based counterparts across various domains \cite{karamcheti2024prismatic,Tong2024Cambrian1AF}. These two types of vision encoders represent two major paradigms: CLIP is trained using image-text contrastive learning, while DINO employs image-only self-supervised learning (Figure~\ref{fig:clip}, top).

This observation raises a fundamental question:
\textbf{Is CLIP’s superior performance primarily due to its language-supervised training objective, or is it simply a result of its significantly larger training dataset?}
Despite the difference in supervision, CLIP models are often trained on datasets that are up to 100 times larger than those used for DINO (Figure~\ref{fig:clip}, bottom). This substantial discrepancy in data scale makes it difficult to disentangle the effects of training objective from those of dataset size.

\begin{table*}[!tb]
\centering
\rowcolors{2}{white}{light-light-gray}
\setlength\tabcolsep{5pt}
\renewcommand{\arraystretch}{1.1}
\small
\begin{tabular}{lccccccc}
\toprule
 & \multicolumn{2}{c}{\textbf{General}} & \multicolumn{3}{c}{\textbf{Fine-grained}} & \multicolumn{2}{c}{\textbf{Robustness}} \\
Models & ImageNet & CIFAR10 & Stanford Cars & Flowers & CUB & ImageNetV2 & CIFAR10.1 \\
\midrule
Official CLIP      & 79.5 & 93.4 & 80.8 & 89.7 & 74.9 & 68.9 & 87.3 \\
Official DINO      & 76.1 & 93.5 & 61.2 & 83.2 & 71.0 & 65.5 & 84.7 \\
Controlled CLIP    & 65.8 & 90.7 & 74.7 & 78.7 & 52.3 & 53.0 & 82.8 \\
Controlled DINO    & 66.4 & 92.1 & 54.1 & 80.7 & 43.0 & 53.5 & 86.0 \\
\bottomrule
\end{tabular}
\label{tab:performance}
\caption{\textbf{Linear probing accuracy (\%) of controlled CLIP and DINO.} Trained under identical settings except for the presence of language supervision in CLIP, both models perform similarly on general and robustness benchmarks. However, CLIP shows significantly higher accuracy on fine-grained classification tasks, highlighting the benefit of language supervision in distinguishing visually similar categories.}
\vspace{-1em}
\end{table*}

To isolate these factors, we conduct a controlled study by training CLIP and DINO vision encoders under identical conditions: the same architecture (ViT-B/16), dataset (a 10M-image subset of DataComp \cite{DBLP:conf/nips/GadreIFHSNMWGZO23}), and training configurations (20 epochs). Notably, the resulting models achieve comparable ImageNet \cite{5206848} linear probing accuracy (CLIP: 65.8\%, DINO: 66.4\%), ensuring a fair basis for comparison.

Using these controlled encoders, we first investigate how language supervision alters the embedding space. We identify and analyze image pairs where CLIP and DINO produce significantly different similarity scores. Our analysis reveals that CLIP is more sensitive to high-level visual semantics—such as object type and embedded text—while DINO is more responsive to low-level visual attributes like colors and styles.

We then integrate these controlled encoders into the LLaVA \cite{Liu2023VisualIT} and train the resulting VLMs under identical settings. Evaluated on 20 VQA benchmarks, LLaVA-CLIP significantly outperforms LLaVA-DINO on text-intensive tasks (e.g., questions involving tables or charts), achieving a 7.5\% performance gain. LLaVA-DINO performs slightly better on some visually grounded tasks but matches LLaVA-CLIP on most others.

To further probe the effect of language supervision, we explore two additional questions: (1) Do different supervision objectives, such as CLIP vs. SigLIP, lead to performance differences? (2) Does using a pre-trained language encoder during CLIP training yield a stronger vision encoder? In both cases, we find the answer to be no.

In summary, our study examines how vision encoders influence the performance of vision-language models. Through carefully controlled experiments, we uncover the representational differences induced by language supervision and their downstream effects, thereby offering the community deeper scientific insights into designing vision-centric vision-language systems.

\section{Training Controlled CLIP and DINO}

In this section, we describe how we train CLIP and DINO under controlled settings, ensuring that the only difference lies in the supervision signal. We then evaluate their performance on various image classification benchmarks.

\paragraph{Experimental setup.}
To isolate the effect of supervision, we align all other factors in CLIP and DINO training. 1) Architecture: We use ViT-B/16 as the backbone for both models; 2) Dataset: We train on a 10M subset of the DataComp \cite{DBLP:conf/nips/GadreIFHSNMWGZO23} image-caption dataset. All images are center-cropped and resized to 224 $\times$ 224. For CLIP, we use the associated captions as language supervision; for DINO, no textual input is provided; 3) Training Configuration: Both models are trained from scratch for 20 epochs using the AdamW optimizer, a learning rate of 1e-3, and cosine learning rate decay. Training is conducted on 4 A100 GPUs over 3 days.

\paragraph{Results.}
After training, we evaluate the encoders using linear probing on standard image classification benchmarks—a widely adopted approach for assessing vision encoder quality. As shown in Table 1, the models perform similarly on general classification tasks such as ImageNet \cite{5206848} and CIFAR-10 \cite{Krizhevsky09learningmultiple}. However, the difference becomes more pronounced on fine-grained classification benchmarks: CLIP significantly outperforms DINO on Stanford Cars \cite{KrauseStarkDengFeiFei_2013} (74.7\% vs. 54.1\%, +20.6\%) and CUB \cite{WahCUB_200_2011} (52.3\% vs. 43.0\%, +9.3\%), despite being trained on the same image data. This suggests that language supervision is especially helpful for tasks requiring detailed semantic distinctions.
For robustness evaluation, performance is comparable between CLIP and DINO, aligning with previous findings \cite{fang2022data}. Overall, these results indicate that while training data scale governs general classification and robustness, language supervision provides substantial benefits for fine-grained recognition tasks where subtle visual differences must be captured.

\section{Embedding Analysis}

\begin{figure}[!tb]
    \centering
    \includegraphics[width=\linewidth]{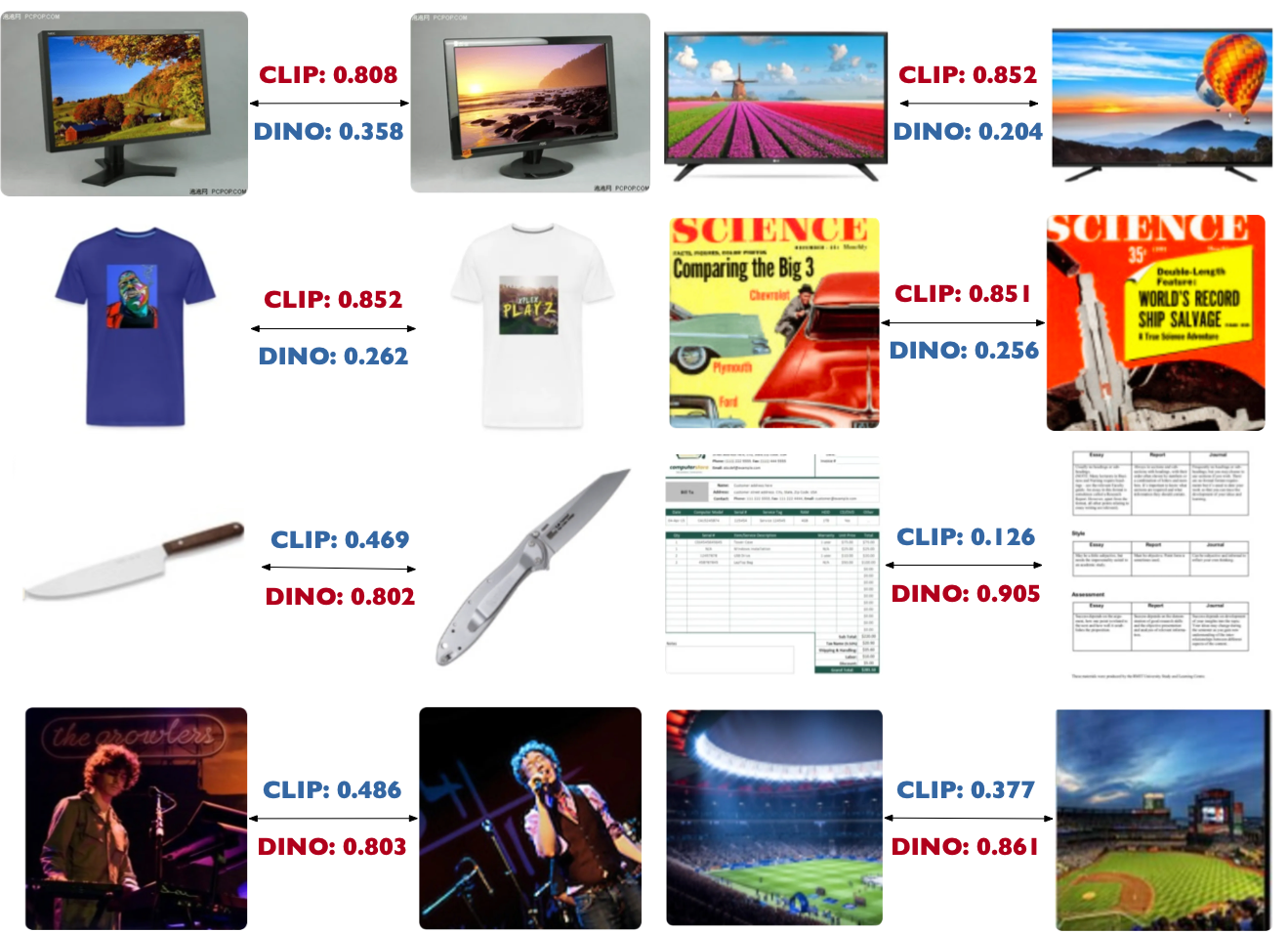}
    \vspace{-1.5em}
    \caption{\textbf{Embedding analysis of CLIP and DINO.} The top two image pairs exhibit high cosine similarity according to CLIP but low similarity under DINO, suggesting that CLIP is more attuned to high-level semantics such as object categories and embedded text. In contrast, the bottom pairs show the opposite pattern, indicating that DINO is more sensitive to low-level features like object colors and visual styles.}
    \label{fig:embedding}
    \vspace{-1em}
\end{figure}

\begin{figure*}[!tb]
    \centering
    \includegraphics[width=\linewidth]{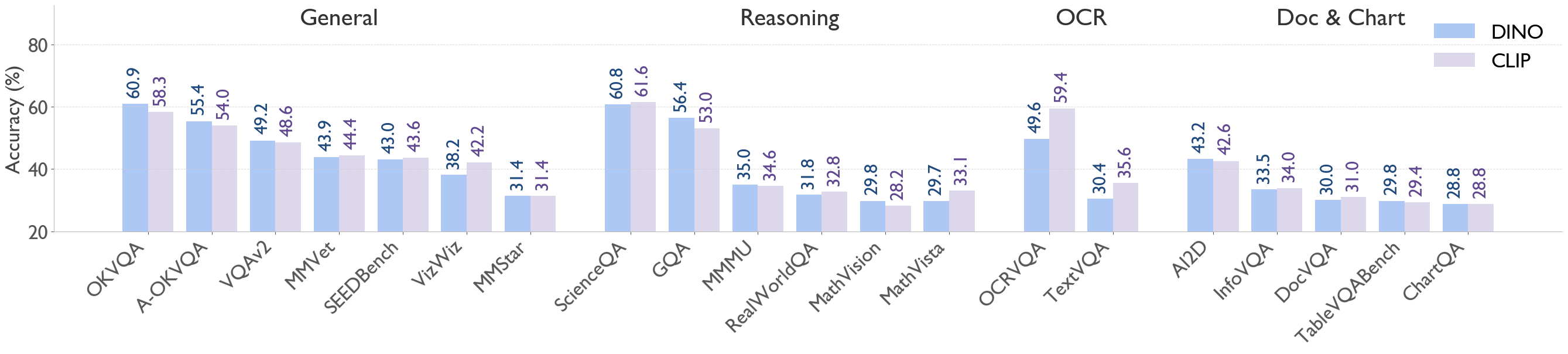}
    \vspace{-2.3em}
    \caption{\textbf{VLM analysis of CLIP and DINO.} We integrate the controlled CLIP and DINO encoders into LLaVA-1.5 and evaluate on 20 subsets of the VMCBench benchmark. Results show that LLaVA-CLIP significantly outperforms LLaVA-DINO on OCR tasks by 7.5\%, while their performance is largely comparable on other tasks.}
    \vspace{-1em}
    \label{fig:vlm performance}
\end{figure*}

To gain deeper insight into how language supervision shapes the embedding space, we conduct a fine-grained embedding analysis comparing CLIP and DINO. Unlike coarse-grained metrics like classification accuracy, this analysis reveals how each model organizes visual information.

\paragraph{Method.}  
Similar to Tong et al.~\cite{Tong2024EyesWS}, we analyze pairs of images in DataComp-10M where CLIP and DINO produce highly divergent similarity scores, revealing systematic differences in representation. Specifically, we identify two types of image pairs:
\begin{align}
    g_1 &= (\texttt{clip\_sim} > 0.8) \land (\texttt{dino\_sim} < 0.5) \nonumber \\
    g_2 &= (\texttt{dino\_sim} > 0.8) \land (\texttt{clip\_sim} < 0.5) \nonumber \label{eq:mask_two}
\end{align}
These selected pairs help isolate cases where the two models disagree in their embeddings.

\paragraph{Results.}
Figure~\ref{fig:embedding} illustrates representative examples for our analysis. CLIP shows strong alignment with high-level semantic features such as object identity and textual content. It consistently groups images by object type or embedded texts, even across variations in visual style or context—suggesting that language supervision enhances semantic abstraction. In contrast, DINO is more sensitive to low-level visual cues like color schemes, and is more invariant to orientation change. We provide a quantitative validation for these observations in Appendix~\ref{sec:quant_analysis}. These findings highlight that CLIP learns embeddings that are more semantically meaningful, while DINO emphasizes visual similarity, likely due to its self-supervised objective.

\section{VLM Analysis}

\label{sec:baseline_comparison}

After training the controlled CLIP and DINO encoders, we incorporate them into the LLaVA-1.5 framework to investigate how vision encoder choice impacts the performance of VLMs.

\paragraph{Experimental Setup.}  
We use LLaVA-1.5 with its vision encoder replaced by either controlled CLIP or DINO. Training consists of pretraining followed by visual instruction tuning. During training, we save checkpoints every 500 steps and evaluate each on VMCBench \cite{Zhang2024automated}—a unified multiple-choice visual question answering benchmark composed of 20 datasets—to simplify evaluation. Since test set labels are not publicly available, we select the best checkpoint based on validation performance and report validation results. All training configurations are kept identical for both CLIP and DINO versions.

\paragraph{Results.}  
Figure~\ref{fig:vlm performance} presents performance across the 20 VMCBench subsets. \textbf{CLIP and DINO perform comparably on most tasks:} On general VQA and reasoning tasks, both encoders yield similar results. For instance, DINO achieves 41.5\% accuracy on reasoning tasks versus CLIP's 41.2\%; for general VQA, CLIP slightly edges out DINO at 46.2\% versus 46.0\%. In document and chart understanding (Doc\&Chart), performance is nearly identical: 33.2\% for CLIP vs. 33.1\% for DINO. These small differences suggest that both encoders are similarly effective in broad VLM tasks. \textbf{CLIP excels in text-intensive visual tasks:} The most notable difference appears in OCR-based benchmarks. On average, LLaVA-CLIP achieves 47.5\% on OCRVQA \cite{8978122} and TextVQA \cite{singh2019towards}, while LLaVA-DINO reaches only 40.0\%, a substantial 7.5 percentage-point gap. This result indicates that language supervision in CLIP enhances its ability to extract and reason over textual content embedded in images—a key capability for text-heavy visual understanding.

\section{Exploring Better Language Supervision}

Given that language supervision (1) improves fine-grained image classification, (2) encourages high-level semantic alignment, and (3) enhances OCR task performance in VLMs, we further explore whether alternative forms of language supervision can yield stronger vision encoders.

\paragraph{Experimental Setup.}  
We explore two directions to improve CLIP's language supervision. First, we replace the standard contrastive loss with the sigmoid-based SigLIP loss to examine whether the training objective affects performance. Second, we substitute the randomly initialized text encoder in CLIP with a frozen, pretrained Vicuna-7B \cite{zheng2023judging} model to assess the value of stronger language priors. After training, we integrate each encoder into the LLaVA-based VLM (as described previously) and evaluate on VMCBench.

\paragraph{Results.}  
As shown in Table~\ref{tab:compare}, neither modification outperforms the baseline CLIP model. Both the SigLIP loss and the pretrained Vicuna-based encoder yield slightly lower average accuracy. These results suggest that while language supervision is critical, the specific form—whether via objective function or pretrained language model—may offer limited additional benefit, consistent with recent observations in the literature~\cite{huang2024llm2clip}.

\begin{table}[!tb]
\centering
\rowcolors{2}{white}{light-light-gray}
\setlength\tabcolsep{5pt}
\renewcommand{\arraystretch}{1.1}
\small
\begin{tabular}{ccc}
\toprule
\textbf{CLIP} & \textbf{SigLIP Loss} & \textbf{Pretrained LM (Vicuna)} \\
\midrule
41.4 & 40.8 & 40.5 \\
\bottomrule
\end{tabular}
\vspace{-0.3em}
\caption{Alternative language supervision objectives or using a pretrained text encoder do not improve CLIP performance when used in vision-language models.}
\label{tab:compare}
\vspace{-1em}
\end{table}

\section{Related Works}
\label{sec:related_works}

\paragraph{Vision-Language Models.}  
Recent years have seen rapid advances in Vision-Language Models (VLMs), with architectures such as LLaVA \cite{Liu2023VisualIT} and Qwen2.5-VL \cite{bai2025qwen25vl} demonstrating increasingly sophisticated multimodal capabilities. These models typically pair a vision encoder with a large language model (LLM), enabling joint reasoning over visual and textual inputs. In this framework, the vision encoder plays a critical role by converting images into representations that can be projected and processed by the LLM. Our work focuses on this vision encoder component, aiming to understand how its training affects downstream VLM performance.

\paragraph{Visual Representation Learning.}  
Visual representation learning mainly followes two paradigms: self-supervised and language-supervised learning. Self-supervised approaches, such as DINO \cite{caron2021emerging} and SimCLR \cite{chen2020simple}, learn representations by predicting relationships between augmented views of the same image. In contrast, language-supervised methods—exemplified by CLIP \cite{DBLP:conf/icml/RadfordKHRGASAM21}, EVA-CLIP \cite{sun2023eva0clip0}, and SigLIP \cite{zhai2023sigmoid}—leverage image-text pairs to align visual and linguistic representations. These two families of methods not only differ in supervision strategy but also in the scale of training data. In this work, we systematically ablate which factor—supervision type or data scale—drives performance gains.

\paragraph{Design Choices in Vision-Language Models.}  
Several studies have investigated how architectural components, data curation strategies, and training configurations affect VLM performance \cite{karamcheti2024prismatic, NEURIPS2024_a0303731, mckinzie2024mm1}. Across these works, CLIP and its variants (e.g., SigLIP) consistently emerge as the most effective vision encoders. However, such findings are typically based on pre-trained models, which differ in supervision objectives, data size, and training setups—making it difficult to isolate the source of performance differences. In contrast, our work trains CLIP and DINO under controlled conditions to isolate the effect of language supervision on vision encoder quality.

\section{Conclusion}

This work conducts a controlled study to disentangle the effects of language supervision and data scale on vision encoder performance in VLMs, offering insights into vision encoder design and its role in effective VLMs.

\section*{Acknowledgments}

S.Y. is a Chan Zuckerberg Biohub — San Francisco Investigator.

\section*{Limitations}

While our study is carefully controlled, it is limited to a 10M-image subset. Scaling these comparisons to billion-image datasets is a crucial next step for fully understanding the interplay between supervision type and data magnitude. A concurrent work addressed this question by scaling DINO and CLIP to 7B parameters on 8B data \cite{fan2025scaling}. Additionally, exploring hybrid approaches that strategically combine self-supervised and language-supervised signals remains a promising direction for advancing vision encoder design.

\bibliography{custom}

\newpage
\appendix
\appendix

\section{Training Curves}
\label{sec:training_details}

\begin{figure}[htbp]
    \centering
    \includegraphics[width=0.9\linewidth]{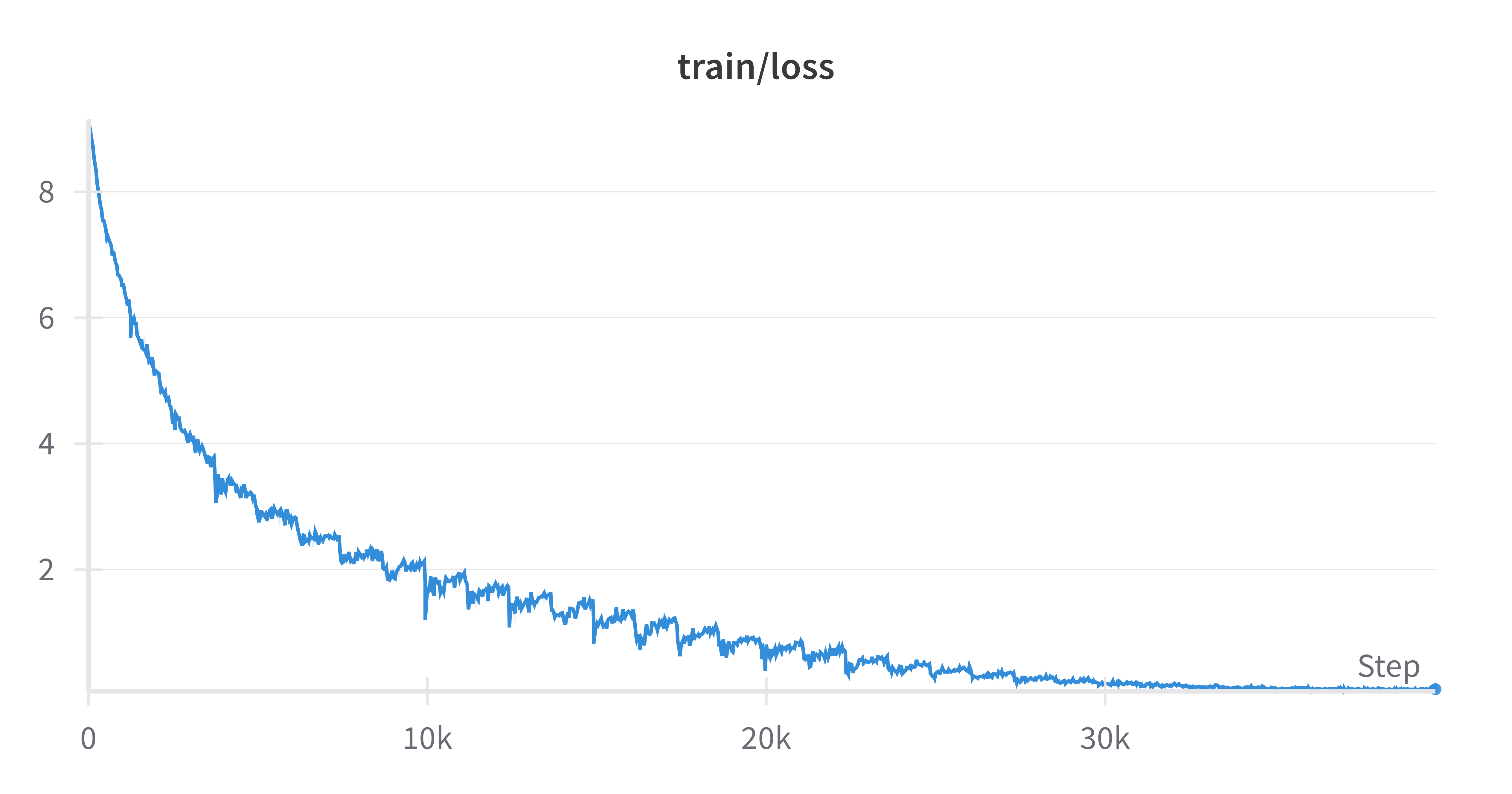}
    \includegraphics[width=0.9\linewidth]{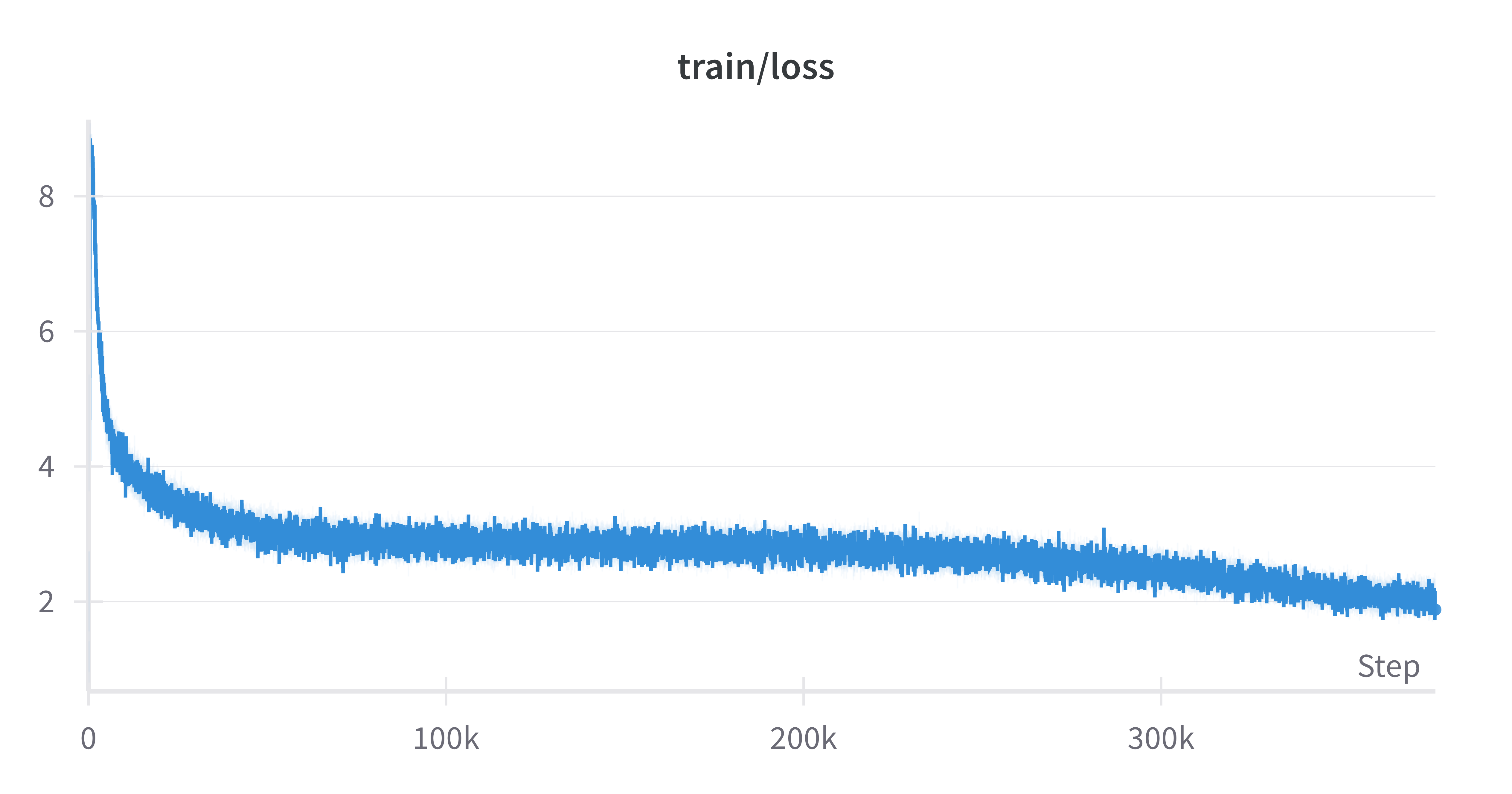}
    \caption{Training loss curves for the controlled CLIP (top) and DINO (bottom) models.}
    \label{fig:training_curve}
\end{figure}

We provide the training loss curves for both CLIP and DINO under the controlled setup. As shown in Figure~\ref{fig:training_curve}, both models converge smoothly within 20 epochs, with no signs of overfitting or instability.

\section{Quantitative Validation of Embedding Space}
\label{sec:quant_analysis}

To provide quantitative support for the claims made in our main embedding analysis (Section 3), we conducted two experiments measuring the cosine similarity within the controlled CLIP and DINO embedding spaces.

\paragraph{Experiment 1: Sensitivity to Semantic Content (Text).}
To test the models' ability to distinguish between high-level semantic concepts, we created a small dataset of images where each image contained a unique alphabet letter or number ('A', 'B', '1', etc.). We then computed the average pairwise cosine similarity between the embeddings of these semantically distinct images.

\textbf{Results:} The average similarity for DINO was 0.877, while for CLIP it was significantly lower at 0.713.

\textbf{Conclusion:} The lower similarity score for CLIP demonstrates that its representations for different semantic symbols are more separable and distinct. This quantitatively confirms that CLIP's embedding space is more structured around the semantic identity of the visual content.

\paragraph{Experiment 2: Sensitivity to Visual Patterns.}
To measure sensitivity to low-level features, we performed a similar analysis on a dataset of images containing simple, repeating visual patterns (e.g., grids, dots, checkers), where semantic content was minimal.

\textbf{Results:} In this case, the trend reversed. The average similarity for DINO was 0.478, while for CLIP it was 0.497.

\textbf{Conclusion:} The lower similarity score for DINO indicates that its representation space separates these low-level visual patterns more effectively. This provides quantitative support for our claim that DINO is more sensitive to visual structure.

Together, these quantitative results align perfectly with our qualitative analysis, providing a robust and comprehensive picture of how language supervision shapes visual representations compared to self-supervision.

\section{Using Qwen2-7B as the LLM Backbone}
\label{sec:qwen}

To further examine the interaction between vision encoders and language models, we evaluate the performance of our controlled CLIP and DINO encoders using Qwen2-7B \cite{yang2024qwen2} as the LLM backbone, in comparison to Vicuna-7B. Results are summarized in Table~\ref{tab:model_combination_scores}.

\paragraph{Improved General VQA Performance with Qwen2-7B.}
When paired with Qwen2-7B, CLIP demonstrates an advantage in general VQA tasks, achieving 57.90\% accuracy compared to DINO’s 54.02\%—a 3.88 percentage point gain. This contrasts with the Vicuna-7B setting, where CLIP and DINO achieved nearly identical results in the same category (46.23\% vs. 46.20\%). These results suggest that Qwen2-7B may better leverage CLIP’s high-level semantic representations for tasks requiring holistic scene understanding.

\begin{table}[htbp]
    \small
        \rowcolors{2}{white}{light-light-gray}
\setlength\tabcolsep{1.5pt}
\renewcommand{\arraystretch}{1.1}
  \centering
  \begin{tabular}{@{}lccccc@{}}
    \toprule
    \textbf{Model} & \textbf{General} & \textbf{Reason} & \textbf{Doc/Chart} & \textbf{OCR} & \textbf{Avg} \\
    \midrule
    CLIP + Vicuna       & 46.23 & 41.17 & 33.15 & 47.50 & 41.44 \\
    DINO + Vicuna       & 46.20 & 41.50 & 33.07 & 40.00 & 40.71 \\
    CLIP + Qwen2        & 57.90 & 47.74 & 40.62 & 51.40 & 49.69 \\
    DINO + Qwen2        & 54.02 & 47.56 & 39.86 & 47.59 & 47.72 \\
    \bottomrule
  \end{tabular}
  \caption{Performance on VMCBench using different vision encoder and LLM backbone combinations. Qwen2-7B leads to stronger performance across most categories, especially when paired with CLIP.}
  \label{tab:model_combination_scores}
\end{table}

\end{document}